\def\year{2018}\relax
\documentclass[letterpaper]{article}  %DO NOT CHANGE THIS
\usepackage{aaai18}  %Required
\usepackage{times}  %Required
\usepackage{helvet}  %Required
\usepackage{courier}  %Required
\usepackage{url}  %Required
\usepackage{graphicx}  %Required
\usepackage{amsmath}
\usepackage{CJKutf8}
\usepackage[table]{xcolor}
\usepackage{array,ragged2e}
\begin{document}
\title{Integrating User and Agent Models: A Deep Task-Oriented Dialogue System}

\author{Weiyan  Wang \and Yuxiang  WU \and Yu  Zhang \and Zhongqi   Lu \and Kaixiang  Mo \and Qiang  Yang  \\
Hong Kong University of Science and Technology, Hong Kong  \\
 \{wwangbc, ywubw,yuzhangcse, zluab,kxmo,qyang\}@cse.ust.hk }

\maketitle
\begin{abstract}
Task-oriented dialogue systems can efficiently serve a large number of customers and relieve people from tedious works. However, existing task-oriented dialogue systems depend on handcrafted actions and states or extra semantic labels, which sometimes degrades user experience despite the intensive human intervention. Moreover, current user simulators have limited expressive ability so that deep reinforcement Seq2Seq models have to rely on selfplay and only work in some special cases. To address those problems, we propose a uSer and Agent Model IntegrAtion (SAMIA) framework inspired by an observation that the roles of the user and agent models are asymmetric. Firstly, this SAMIA framework model the user model as a Seq2Seq learning problem instead of ranking or designing rules. Then the built user model is used as a leverage to train the agent model by deep reinforcement learning. In the test phase, the output of the agent model is filtered by the user model to enhance the stability and robustness. Experiments on a real-world coffee ordering dataset verify the effectiveness of the proposed SAMIA framework.
\end{abstract}

\section{Introduction}

Task-oriented dialogue systems aim to help people to complete some practical tasks such as food ordering, booking, and information query. Unlike chit-chat dialogue systems in the open domain, they can provide some specific services to a large number of customers efficiently and relieve people from repeated and tedious works. To complete such practical tasks, a task-oriented dialogue system usually needs to deal with multi-turn dialogues, which requires to properly consider context and logics among the different turns. Table~\ref{tab:dl_example} shows a real-world coffee ordering conversation with the dialogue log and order information.

\begin{table}[!ht]
\centering
\caption{Example of multi-turn task-oriented dialogue for ordering coffee, which is translated from Chinese to English by authors.}
\label{tab:dl_example}
\begin{CJK*}{UTF8}{gbsn}
\begin{tabular}{|p{45mm}|p{29mm}|}
\hline
\Centering Conversation Log &  \Centering Order Information \\
\hline

{
U: Can you help me to order starbucks?\newline
A: What would you like to drink? Where to send?\newline
U: No.$<$NUM$>$ Building, Software park, Zhongguancun\newline
A: What would you like to drink?\newline
U: Caramel Macchiato, the medium size\newline
A: Hot?\newline
U: Hot\newline
A: Your coffee order has been confirmed,please click to view details and pay \newline
U: How long will it take?\newline
A: About $<$NUM$>$ minutes
} &
{
\textbf{Address:} \newline No. $<$NUM$>$  Building, Software park, Zhongguancun\newline\newline
\textbf{Taste:}\newline Caramel Macchiato\newline\newline
\textbf{Size:} Medium \newline\newline
\textbf{Temperature:} Hot \newline\newline}  \\
\hline
\end{tabular}
\clearpage\end{CJK*}
\end{table}

Traditionally, task-oriented dialogue systems are modeled as a reinforcement learning problem \cite{levin1997learning,young2013pomdp} in which the reward is given depending on whether the task is accomplished. The states and response templates need to be carefully handcrafted for each specific task. They are usually bootstrapped by agenda-based or event-based user simulators \cite{schatzmann2007statistical,asri2016sequence}, which again require domain knowledge as well as careful engineering. Therefore, even with such intensive handcrafting they cannot deal with varieties of natural language and generate proper responses in many practical situations, resulting in degrading user experiences. 

Recently,  sequence-to-sequence (Seq2Seq) models \cite{sutskever2014sequence,bahdanau2014neural,sordoni2015hierarchical} appear to be a promising alternative approach with the potential to understand and generate natural language sentences. By following the idea of statistical machine translation (SMT), some works based on supervised learning \cite{DBLP:conf/emnlp/WenGMSVY15,wen2016network} converts multi-turn conversations into one-turn SMT problems by adding additional information. They generated based on the input utterance as well as the additional semantic label information such as users' intentions and belief states. Some other works based deep reinforcement learning \cite{li2015diversity,DBLP:journals/corr/LewisYDPB17} hold an assumption of symmetric roles in the dialogue and depend on selfplay for training, which is not appliable for customer and client's asymmetric roles in most task-oriented dialogues.

In a word, these two approaches suffer from some drawbacks when modeling multi-turn task-oriented dialogue systems. For example, they require additional information like handcrafted states and actions required by traditional MDP/POMDP approaches and extra supervised information for supervised learning based Seq2Seq approach. Moreover, existing user models require intensive human interventions and have too limited expressive ability to train a seq2seq agent model based on deep reinforcement learning. All previous attempts on dialogue system based on deep reinforcement learning have to rely on the selfplay and limit themselves in a small scope of symmetric dialogues like chitchat or negotiation.

In order to overcome these drawbacks, we propose a uSer and Agent Model IntegrAtion (SAMIA) framework to model both the user and agent simultaneously without any handcrafting. The reason to do that is the observation that roles are asymmetric in the task-oriented dialogue, in which one part is easy while the other part is more complex . As illustrated in Table~\ref{tab:dl_example} , the user usually needs to answer the agent's questions and occasionally ask some one-turn quick questions. However, the agent model needs to identify and process the user's demand in a multi-turn dialogue. Again taking the conversation in Table~\ref{tab:dl_example} as an example, the agent needs to identify user's detailed demands including the taste, cup size, temperature and delivery address through the entire conversation. Therefore, the modeling of the agent and user in task-oriented dialogue systems is different. Specifically, the SAMIA framework novelly models the user as a Seq2Seq model with the potential to understand and generate any natural language, which is learned from one-turn dialogues. Since the task-oriented dialogue system needs to collect user demands each of which is defined as a slot, we use the collection of missing slot information as a reward and hence formulate the agent modeling as a deep reinforcement learning problem to maximizing the expected reward. During its training process, agent model uses user model as a leverage and take the output of the user model as its input to generate the natural and proper response. In the test phase, we employ the user model to predict the user's responses for filtering responses candidates from beam search of the the agent model to enhance the robustness. %The effectiveness of the proposed SAMIA framework has been verified by experiments on a real-world coffee ordering dataset.

The main contribution of this paper is two-fold: (1) The proposed SAMIA framework explicitly and novelty models the user as a Seq2Seq problem resulting in much more expressive ability; (2) Taking the user model as a leverage, the proposed SAMIA framework has extended the scope to the asymmetric situation and modeled agent as a deep reinforcement learning problem. Finally, we conduct experiments on a real-world coffee ordering dataset to show the effectiveness of the SAMIA framework.
\section{Related Works}

Both MDP \cite{levin1997learning,walker2003trainable} and POMDP \cite{young2010hidden,young2013pomdp} are widely used to model the state transition for multi-turn dialogues in task-oriented dialogue systems. Although some works\cite{DBLP:conf/acl/WilliamsAZ17} has even employed end-to-end learning method, these works often still rely on handcrafted templates of states and actions \cite{cuayahuitl2016simpleds} or reward designed for the specific domain. For better performance, some auxiliary information has been utilized and for example, an online active model \cite{su2016line} has been proposed to predict the user's satisfaction rate as the reward for the system, while the other components remain being carefully handcrafted. Furthermore, event-sequence to event-sequence learning based \cite{asri2016sequence} and agenda-based \cite{schatzmann2007statistical} user simulators, which  model discrete user state transitions, have been introduced to boost the performance of the task-oriented system. However, the handcrafted events, templates or rules not only limit the capacity and scope of user simulators in practice but also require intensive labors, so they cannot help train a deep reinforcement dialogue system that needs to understand and generate any valid natural language.

Recently, the Seq2Seq learning \cite{sutskever2014sequence} with the encoder-decoder structure network has achieved remarkable successes in many NLP tasks including translation \cite{bahdanau2014neural}, summarization \cite{rush2015neural}, query prediction \cite{sordoni2015hierarchical} as well as dialogue systems \cite{shang2015neural,serban2015hierarchical}. However, most of these works are based on supervised learning and follow the idea that formulates the response generation as an SMT problem \cite{ritter2011data} and can only generate one-round dialogues. Moreover, a new diversity-promoting objective function \cite{li2015diversity}  and new rewards \cite{li2016deep} are defined to increase the diversity of the generated responses by Seq2Seq models. There are also several attempts have been made to build deep reinforcement Seq2Seq dialogue system using selfplay \cite{li2016deep,DBLP:journals/corr/LewisYDPB17}, but they hold a strong assumption on symmetric roles. All the aforementioned works focus on dialogues between symmetric roles like open-domain chit-chat or negotiation other than common task-oriented dialogue that happens between total different roles and usually processes practical tasks in multi-turn dialogues. 

To build task-oriented dialogue systems, Seq2Seq approach employs extra supervised labels such as users' intentions as well as the selected information slots \cite{DBLP:conf/emnlp/WenGMSVY15,wen2016network} to map dialogue logs between the user and agent in previous turns to the next response. But it does not learn the state representations or incorporates the goal of a task into the modeling process and it additionally needs intensive human efforts to label the detailed information. By storing dialogue logs in a memory, the memory network \cite{bordes2016learning} can compute the ranking of all the possible responses in a large set but this information retrieval approach may fail when labeled data are scarce. Dhingra et al. \cite{dhingra2016end} claim to have developed the first fully end-to-end differentiable model of a multi-turn information-providing dialogue agent, but they mainly rely on a stochastic user query simulator to bootstrap and mainly focus on the soft query of knowledge base other than the response generation in task-oriented dialogue systems.

\section{The SAMIA Framework}

\begin{figure}[!ht]
\centering
\includegraphics[scale=0.4]{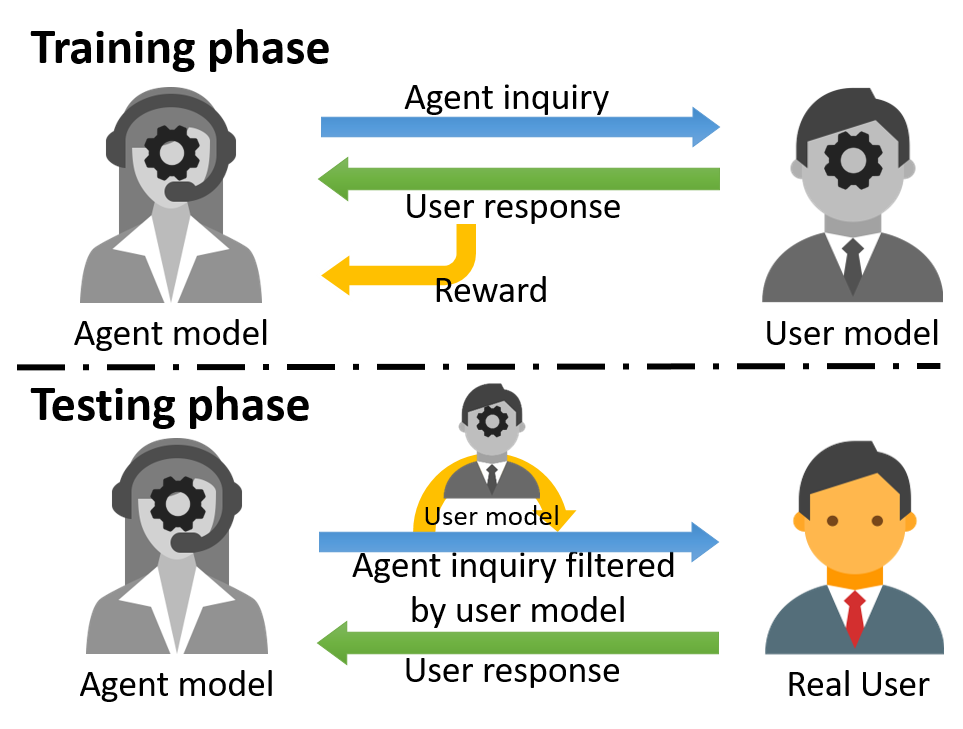}
\caption{The framework}
\label{fig:framework}
\end{figure}

To begin with, we give an overview of the whole \mbox{SAMIA} framework. SAMIA model both the user and agent as seq2seq learning problems with supervised learning and reinforcement learning respectively. Since selfplay is not appliable for this application with asymmetry roles, it is not trivial to model the user as a seq2seq model that can read in any natural language and generate varieties of fluent sentences. As shown in Figure \ref{fig:framework}, the user and agent models work together to do the dialogue simulations and replies from by user model can form the reward to guide the reinforcement learning in training phase. In the test phase, the user model helps the agent model look ahead to filter response candidates and enhance the robustness. 

We take two steps to build the whole framework. Firstly, we start with the easy task of building a user model. The user model is built from the raw logs by taking one-turn agent utterance as the input to predict the user response. Then the agent model takes the user model as a leverage to learn the logic of the multi-turn dialogue via the deep reinforcement learning. Specifically, The agent model is built up in the following two steps: (1) using the supervised learning to pre-train a policy network to learn to generate fluent responses and answer some one-turn quick questions. (2) using the reinforcement learning to update the policy network and learn the logic in the multi-turn dialogue with the user model’s prediction of the next utterance to estimate the reward of current output.In the following sections, we will present the details of the proposed SAMIA framework.

\subsection{The User Model}\label{sec:user_model}

In the task-oriented dialogues, a user usually firstly shows the intention to the agent and then answers the agent’s questions one by one to specify the demand. Occasionally the user asks the agent some one-turn quick questions such as ``how soon will the delivery arrive''. In other words, the user basically gives response only to the question by the agent in one turn. So we can build the user model under a proper assumption that a user only needs to take one-turn dialogue into consideration to give a respond, which leaves the logic of multi-turn dialogues handled by the agent model. This assumption will be verified by experimental results.

\begin{figure}[h]
\centering
\includegraphics[scale=0.6]{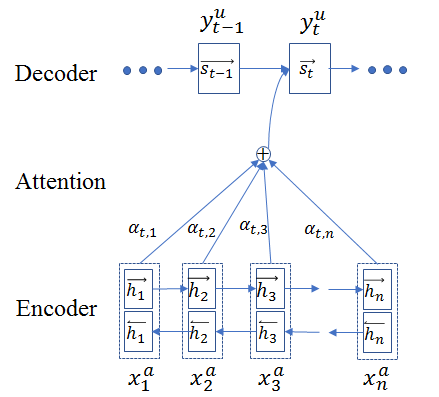}
\caption{The network structure: encoder-decoder structure with the attention mechanism}
\label{fig:seq2seq}
\end{figure}

Specifically, the user is modeled as a Seq2Seq problem based on the bidirectional LSTM network with the encoder-decoder structure as well as the attention mechanism \cite{bahdanau2014neural} as shown in Figure \ref{fig:seq2seq}. By directly learning from the raw dialogue logs, the network takes the agent utterance $X^a:x_1^a,x_2^a,...,x_n^a$ as the input sequence and takes corresponding user utterance $Y^u:y_1^u,y_2^u,...,y_m^u$ as the target sequence. In this model, the conditional probability for each $y_i^u$ is defined as:
\begin{equation}
\label{eqn:nmt}
p(y_i^u|y_1^u,...,y_{i-1}^u,X^a)=g(y_{i-1}^u,s_i,c_i)
\end{equation}
where the context vector $c_i$ is generated by a weighted sum of encoder hidden states $h_j:[\underset{h_j}{\leftarrow},\underset{h_j}{\rightarrow}]$ as $c_i = \sum_{j=1}^{T}\alpha_{ij} h_j$ with the attention weight $\alpha_{ij}$ computed as $\alpha_{ij}=\frac{\exp(e_{ij})}{\sum_{k=1}^{T}\exp (e_{ik})}$ where $e_{ij} = a(s_{i-1}, h_j)$ is to estimate the relatedness between $s_{i-1}$ and $h_j$, and $s_i = f(s_{i-1},y_{i-1}^u,c_i)$ denotes the decoder's \mbox{RNN} hidden state corresponding to index $i$.

Theoretically, if given enough and proper data this seq2seq user model have the potential to be able to understand any natural language and generate any possible natural user response. It is very important this user model learn to be expressive enough so that it can help build a seq2seq deep reinforcement learning agent that understand and generate any natural sentences as well.
\subsection{The Agent Model}

Besides understanding natural language and generating fluent responses, the key to building the agent model is to track the states and logic in the multi-turn dialogue. Namely, the agent model should consider both sequences of sentences in the dialogue and sequences of words to form natural language. Therefore, we reuse the same network structure of user model in Figure \ref{fig:seq2seq}, but we further employ reinforcement learning to model the multi-turn task-oriented dialogue system where supervised learning is only used to do pre-training. The built user model helps establish the environment in reinforcement learning to define the reward.

\subsubsection{Supervised Learning: Pre-train}

\begin{figure}[h]
\centering
\includegraphics[scale=0.45]{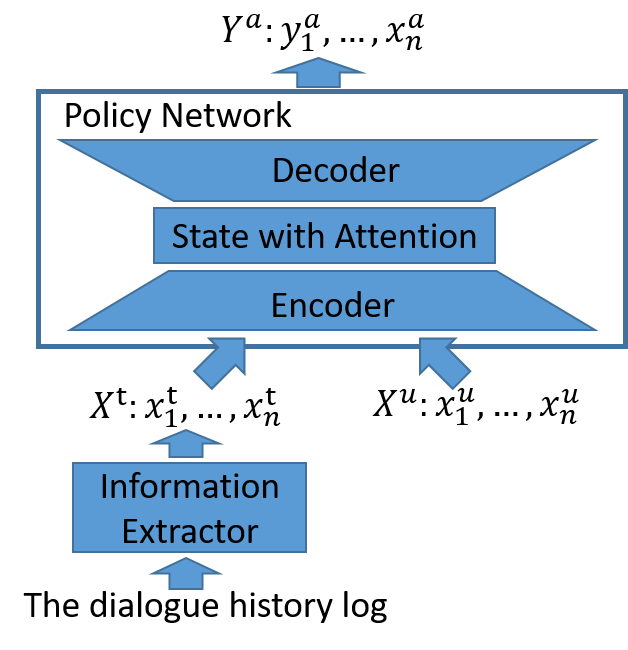}
\caption{The network structure for the supervised learning as pre-training in the agent model.}
\label{fig:taginput}
\end{figure}

Firstly we pre-train the agent model by supervised learning on the raw utterances in the current turn and the tags about the information slots which have been captured in the previous conversation. Despite the same encoder-decoder structure with attention adopted from the user model, the agent model swaps the input and output sequences, and it also takes the tag of filled information slots as an input which is extracted from dialogue in previous turns by pattern matching with the order information in ground truth. As shown in Figure \ref{fig:taginput}, the input includes the filled tags $X^t:x_1^t,x_2^t,..,x_n^t$ in previous turns  as well as the current user's utterance $X^u:x_1^u,x_2^u,...,x_m^u$, and the target output is the agent's utterance $Y^a:y_1^a,y_2^a,...,y_l^a$. The conditional probability of each $y_i^a$ is defined as:
\begin{equation}
\label{eqn:agent nmt}
p(y_i^a|y_1^a,...,y_{i-1}^a,X^t,X^u)=g(y_{i-1}^a,s_i,c_i),
\end{equation}
where function $g(\cdot)$ is similar to Eq. (\ref{eqn:nmt}) except that the context vector $c_i$ is generated from hidden states $s_i$ of both tags $X^t$ and the utterance $X^u$. Although the pre-train model is very similar to previous works based on supervised learning\cite{DBLP:conf/emnlp/WenGMSVY15,wen2016network}, we do not follow them to concatenate tag information and context information but choose to learn proper attention weights on the tags to make better use of the attention mechanism. Additionally, any other semantic intention label is not required in the SAMIA framework.

 Although it does not work very well to learn the multi-turn logic, it helps to generate fluent and proper sentences and to map one-turn quick questions by the user to brief responses. The pre-train will work as a better initialized point than random one and narrow down the search space in the reinforcement learning.

\subsubsection{Reinforcement Learning}

After pre-trained by supervised learning, the parameters of the policy network are initialized in a region that can generate fluent and proper sentences word by word, but it does not explicitly consider the logic in the whole dialogue. In order to achieve better performance, we employ the policy gradient method \cite{williams1992simple,ranzato2015sequence} to update the policy network. The loss function of the policy network at this stage is defined as the negative expected reward:
\begin{eqnarray}
\label{eqn:loss reinforce}
\nonumber
L_{\theta} &=& -\mathbf{E}_{A_{i}\sim p_{\theta}} [r(A_i|X^t,X^u)] \\
&=& -\sum_1^T p_{\theta}(A_i|X^t,X^u)r(A_i|X^t,X^u),
\end{eqnarray}
where $r(\cdot)$ stands for the reward function and $A_i$ denotes the generated sentence consisting of a sequence of words $a^i_1,a^i_2,...,a^i_T$. At time step $t$, the policy network choose one word $a^i_t$ from vocabulary to output as the action, so Eq. (\ref{eqn:loss reinforce}) can be rewritten as
\begin{eqnarray}
\label{eqn:micro loss reinforce}
\nonumber
L_{\theta} &=& -\sum_{i}p_{\theta}(a^i_1,...,a^i_T|X^t,X^u)r(a^i_1,...,a^i_T|X^t,X^u) \\
&=& -\mathbf{E}_{a^i_1,...,a^i_T\sim p_{\theta}}[r(a^i_1,...,a^i_T|X^t,X^u)].
\end{eqnarray}
Since the sentence $A_i: a^i_1,...,a^i_T $  is the output of the network, we can back-propagate the gradients of the reward on each single word in the sequence $A_i:a^i_1,...,a^i_t$ to update the policy network as derived in \cite{williams1992simple} to increase or decrease the generation probability of $A_i$:
\begin{equation}
\frac{\partial L_{\theta}}{\partial \theta}= \sum_t  \frac{\partial L_{\theta} }{\partial a^i_t }\frac{\partial a^i_t}{\partial \theta}.
\end{equation}
However, the reward function is not well defined in dialogue systems. Unlike the BLEU score in the MIXER method \cite{ranzato2015sequence}, whether $A_i$ is helpful to accomplish the task cannot be determined by a deterministic algorithm based on dialogue logs only. What is more, the selfplay does not work in this application since the roles in the dialogue are not symmetry. Therefore, simulations between the agent and user models are employed to generate the reward, which will be discussed in the next sub-section.
\begin{figure}[h]
\centering
\includegraphics[scale=0.45]{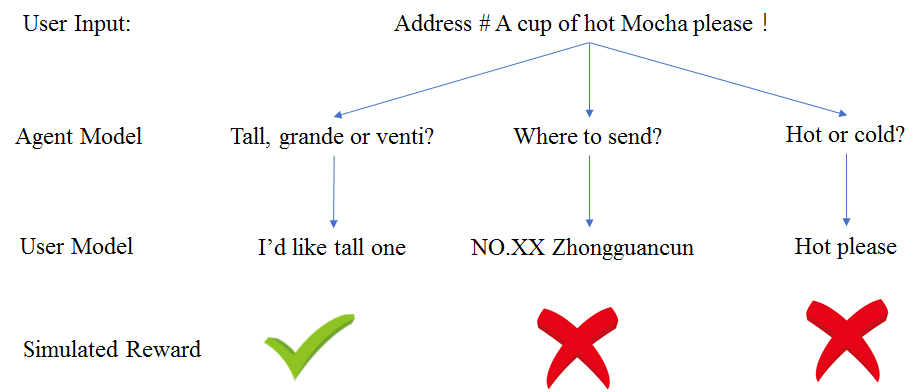}
\caption{Simulation between user and agent models: the left one gets reward because of new information about size, the middle one fails because it gets the same information with the tag, and the right one fails because it gets repeated information about temperature}
\label{fig:simulation}
\end{figure}
\subsubsection{Simulation between Agent and User Models}
In the simulation, the aforementioned user model is used as a leverage to help generate the corresponding reward to the action sequences $A_i$ as illustrated in Figure \ref{fig:simulation}.  Specifically, in the simulations the policy network is firstly fed with the user utterance $X_i^u$ in current turn and the tags of information slots $X_i^t$ obtained from dialogue logs. Then we randomly sample several responses $A_i^{(1)},\ldots,A_i^{(n)}$ as the output candidates. Then the user model takes each sample $A_i^{(j)}$ as input and use the beam search to generate user response denoted by $O_i^{(j)}$. Then based on the utterance pair of the sampled output of the policy network and the output of the user model, we can judge whether there is a vacant information slot to be filled and so the reward assigned to the action $A_i^{(j)}:a_i^{(j,1)},...,a_i^{(j,t)}$ can be defined as the following equation:
\begin{equation}
r(A_i^{(j)}|X^t,X^u) = I(A_i^{(j)},O_i^{(j)}|X^t)-\bar{r},
\end{equation}
where $I(\cdot)$ is an indicator function to judge whether there is a vacant information slot filled by the pair $(A_i^{(j)},O_i^{(j)})$ and $\bar{r}$ denotes the average reward. Here the indicator function $I(\cdot)$ can be simply a pattern matching rule for simple cases or a classifier in a sophisticated case. To be more general, the simulation can even last for several turns between the agent and user models and correspondingly $r(A_i^{(j)}|X^t,X^u)$ is the sum of values of the indicator function in these turns.

\subsubsection{Test Phase}

After training with the reinforcement learning, we can randomly sample the output of the policy network to generate the response, but it may not be stable enough to guarantee the service quality. Therefore, we do the simulation described above with response candidates in the test phase to improve the stability. Specifically, we employ the beam search to generate the candidate set of 20 sampled agent responses and then only the top 5 agent responses will be fed to the user model. With the expected corresponding user response generated by the user model, it is easy to judge how much missing information is and so the candidate responses can be re-ranked by the expected quantity of new information. Finally, our model outputs the top candidate $A_i$ with the largest reward assigned or simply the top-1 one if all candidates do not get any reward.

\subsection{Implementation Details}

In our implementation, the agent and user model share exactly the same network structure with 1024-dimensional hidden state and 256-dimensional  word embedding. Both encoder and decoder only have a single layer of LSTM cells whose maximum length is 50. The vocabulary size of the dictionary is 18408.

In reinforcement learning, we find that only one-turn dialogue need to be simulated in our task to achieve satisfactory performance. Moreover, $\bar{r}$ is set to a zero function, only one response candidate $A_i$ is randomly sampled from the policy network, and the best one from top 20 beam search candidates of the user model corresponding to $A_i$ is used to generate the reward. Specifically, the reward score is 1 if the output of the user model contains new information and otherwise 0.

Supervised learning and reinforcement learning are jointly training to speed up the training process. That is, we do not begin the reinforcement learning until the supervised learning has reached the convergence but start the reinforcement learning with the simulation once the policy network begins to generate fluent sentences. In this way, the updates in both supervised learning and the reinforcement learning can be done on policy network alternatively. One benefit is that we can parallelize supervised learning and the simulations to save time. This joint training strategy also helps the policy network maintain the ability to generate fluent sentences and to answer some quick questions like ``how to pay'' or ``when to arrive'' in case that it is over reinforced, and tries to collect information for a reward even when answering quick questions.

\section{Experiment}

In this section, we empirically evaluate the proposed SAMIA framework.

\subsection{Dataset}

We use a real-world coffee ordering conversation dataset provided by an online O2O company. An example of the dialogues in this dataset is illustrated in Table \ref{tab:dl_example}. In total, there are 31,567 conversation sessions consisting of 142,412 conversation pairs in the dataset. More information about the dataset is shown in Table \ref{tab:dataset}.

\begin{table}[!ht]
\centering
\caption{Statistics of the coffee ordering dataset.}
\label{tab:dataset}
\begin{tabular}{c c c c}
\hline
& \#sessions &\#pairs & proportion \\\hline
 \textbf{training} & 25,264 &120,297 & {80}\% \\
 \textbf{validation} &3,166 &15,107& 10\% \\
 \textbf{test} &3,137 & 14,305 & 10\% \\\hline
 \textbf{total} & 31,567 & 149,709 & 100\%\\
\hline
\end{tabular}
\
\end{table}

Contrary to the DSTC dataset \cite{henderson2014second} with clean data and rich supervised labels including the shallow representation of the utterance semantics, ontology of the feasible set of calls, tracker outputs and the tags of the information slots, the dataset used in our experiment has only the tags of filled information slots extracted by pattern matching between dialogue log and final order information. Moreover, this dataset is also noisy and has a variance in the use of language, in spite of the larger size than that of the DSTC dataset. %The conversations we used are from coffee ordering domain other than restaurant inquiry, and the language is Chinese instead of English.

\subsection{Experiments on User Model}

\begin{figure}[!htb]
\centering
\includegraphics[width=0.5\textwidth]{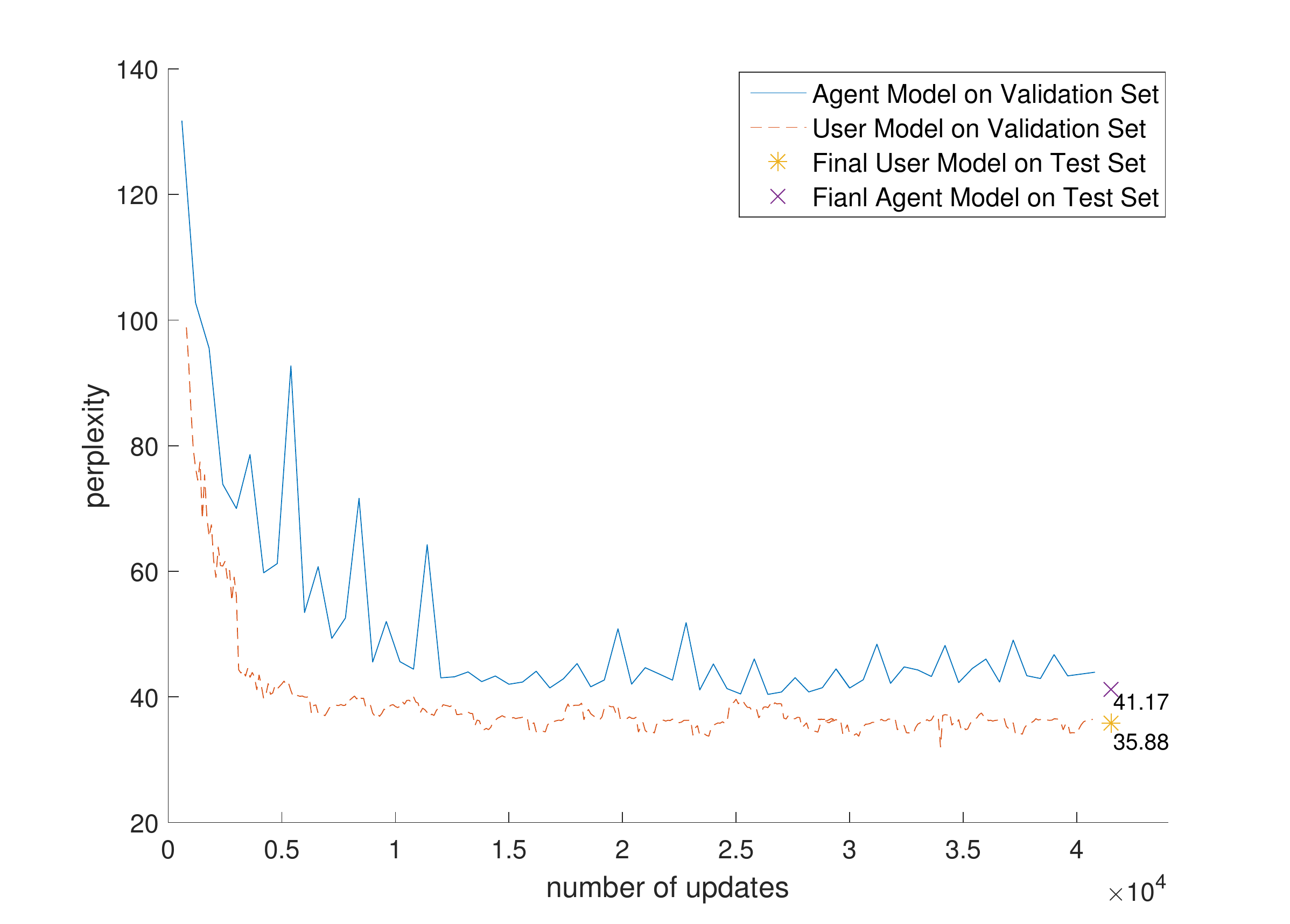}
\caption{The comparison of the perplexities between the user and agent models.}
\label{fig:user_vs_agent}
\end{figure}

To verify our assumption that the user model mainly considers one-turn dialogues but the agent model cannot, an experiment is conducted by building the one-turn supervised learning models described in Section \ref{sec:user_model} for both the user and agent models. Figure \ref{fig:user_vs_agent} shows perplexities of both the user and agent models on the validation set during the training process and their perplexities on the test set. According to Figure \ref{fig:user_vs_agent}, during the training process, the perplexity of the user model is always lower than that of the agent model and it is stably decreasing with the updates going on, while the agent model has more fluctuations. Most importantly, the perplexity of the trained user model on the test set (i.e.,  35.89) is much lower than that of the agent model (i.e.,  41.17).

As cases shown in the Table \ref{tab:sl_case}, both the user and agent models can generate fluent sentences. The agent model can also answer some one-turn quick questions such as ``how to pay" or ``when to deliver". But it fails to logically ask for the user's specific demands and sometimes repeatedly ask the same question or place orders too early. One reason for that is that for the same user utterance, there can be multiple feasible responses of the agent, which makes it difficult for training. On the other hand, the user model mainly completes an easier task to directly answer questions asked by the agent in most cases, hence it performs much better.

\begin{table}[!ht]
\centering
\caption{The case study of the user and agent models when using a seq2seq model for training.}
\label{tab:sl_case}
\begin{tabular}{|p{35mm}|p{40mm}|}
\hline
 \Centering \textbf{User Model} &  \Centering \textbf{Agent Model} \\\hline
{
\textbf{I:} What kind of coffee would you like to drink? \newline
\textbf{O:} Latte.
}
&
{
\textbf{I:} I'd like a cup of coffee\newline
\textbf{O:} What coffee would you like?And  Where to deliver?
}
\\\hline
{
\textbf{I:} Cold or hot? \newline
\textbf{O:} Hot.
}
&{
\textbf{I:} Hot tall Latte\newline
\textbf{O:} Your order will be placed very soon. Link for payment will be sent shortly.
}
\\\hline
{
\textbf{I:} Tall size? \newline
\textbf{O:} Yeah.
}
&{
\textbf{I:} Cold  (or tall size) please\newline
\textbf{O:} Your order will be placed very soon. Link for payment will be sent shortly.
}
\\\hline
{
\textbf{I:} Where to deliver?\newline
\textbf{O:} NO. $<$NUM$>$ Middle Dongfeng  Road, Yuexiu, Guangzhou
}
&{
\textbf{I:} Sent to the Software Park, Zhongguancun \newline
\textbf{O:} What coffee  would you like to drink?
}
\\\hline
{
\textbf{I:} Your order has been placed. Please pay by clicking\newline
\textbf{O:} OK, thanks.
}
&
{\textbf{I:} How long will it take?\newline
\textbf{O:} Usually about one hour .
}
\\\hline
\end{tabular}

\end{table}

\begin{table*}[!ht]
\centering
\caption{An example for dialogues by different models translated from Chinese to English by authors.}\label{tab:example}
\begin{CJK*}{UTF8}{gbsn}
\begin{tabular}{p{35mm}|p{35mm}|p{39mm}|p{46mm}}
\hline
 \Centering {\textbf{SLNT}} &  \Centering {\textbf{SLT}} & \Centering {\textbf{SAMIA-A}} &  \Centering {\textbf{SAMIA}} \\
\hline

{U: I want to order coffee\newline
A: What coffee do you want? Where to deliver?\newline
U: Latte. Y company, number X of Tonglinge road.\newline
A: Tall, grande or venti? Hot or cold?\newline
U: Grande\newline
A: Your order will be placed very soon. Link for payment will be sent shortly.\newline
U: How long will it take?\newline
A: About an hour.} &
{U: I want to order coffee\newline
A: What coffee do you want? Where to deliver?\newline
U: Latte. Y company, number X of Tonglinge road.\newline
A: Ok, your order (grande hot latte) has been confirmed, please click to check details and pay.\newline
U: How long will it take?\newline
A: About an hour.} &
{U: I want to order coffee\newline
A: What coffee do you want? Where to deliver?\newline
U: Latte. Y company, number X of Tonglinge road.\newline
A: Tall, grande or venti? Hot or cold?\newline
U: Grande please\newline
A: OK!\newline
U: Hot\newline
A: Your order will be placed very soon. Link for payment will be sent shortly.\newline
U: How long will it take?\newline
A: About X minutes.} &
{U: I want to order coffee\newline
A: What coffee do you want? Where to deliver?\newline
U: Latte. Y company, number X of Tonglinge road.\newline
A: Tall, grande or venti? Hot or cold?\newline
U: Grande\newline
A: Hot or cold?\newline
U: Hot\newline
A: Your order (grande hot latte) has been confirmed, please click to pay.\newline
U: How long will it take?\newline
A: About X hour(s).}\\
\hline
\end{tabular}
\clearpage\end{CJK*}
%\caption{\label{tab:example}Example dialogue by different models. The second row is the English translation of the first row. Numbers are represented by ``X'' and organizations are anonymized into ``Y''.}
\end{table*}

\subsection{Experiments on Multi-Turn Dialogue Generation}

In this section, we evaluate the proposed SAMIA framework to generate responses in multi-turn dialogues. We handcraft a rule-based reward generator with fixed patterns of questions for the criterion-referenced evaluation. The \mbox{SAMIA} framework is compared with the following  baseline models:
\begin{itemize}
\item SL without tag (SLNT): the supervised learning model trained on user and agent utterances only.
\item SL with tag (SLT): the supervised learning model trained on user and agent utterances as well as extra supervised labels on the tags of information slots.
\item SAMIA agent only (SAMIA-A): only the trained agent model of SAMIA without the user model.
\end{itemize}

\subsubsection{Evaluation Metrics}

Since there is no golden evaluation for the dialogue system, both objective and subjective methods are used to evaluate. Specifically, we adopt the following 3 evaluation metrics:
\begin{itemize}
\item Average Reward: With the rule-based reward generator, it quantitatively measures how much information is collected by the lowest ppl response in the top 5 beam search candidates of each model.
\item Normalized Discounted Cumulative Gain (nDCG): As a commonly used metric in the information retrieval, it evaluates the performance of all the candidates generated in the beam search. The candidate responses by the agent model are treated as documents and their corresponding rewards as gains. The Discounted Cumulative Gain (DCG) accumulated at position $p$ is expressed by DCG$_p=r_1 + \sum_{i=2}^{p}\frac{r_i}{\log_2{i}}$, where $r_i$ is the reward of $i$-th response.  nDCG$_p$, the nDCG accumulated at position $p$, is computed by normalizing with the maximum.
\item Human Evaluation: 4 persons are invited to evaluate all models on $10$ sessions sampled from test dataset. Three aspects are considered for the evaluation including \emph{logic}, \emph{language} and \emph{overall quality}. Each aspect is marked on a scale from $0$ to $2$ where score $0$ means `not suitable', $1$ `somewhat suitable', and $2$ `very suitable'.
\end{itemize}

\subsubsection{Experimental Results}

\begin{figure}[!htb]
\centering
\includegraphics[scale=0.5]{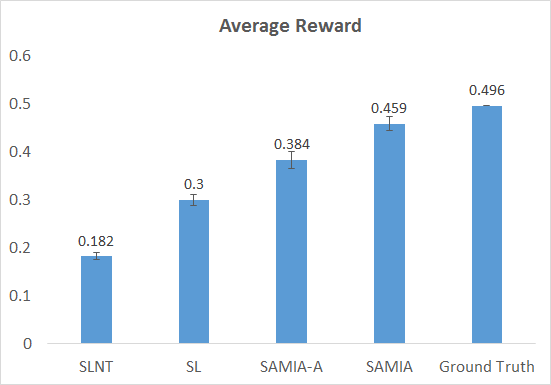}
\caption{Average rewards of different models.}
\label{fig:avg_reward}
\end{figure}
The average rewards and standard error bar of different models on test repeated  five times are shown in Figure \ref{fig:avg_reward}. In this criterion-referenced evaluation, SAMIA-A alone significantly outperforms both the SLNT and SLT models. Compared with SLNT, the SLT has a slight improvement in the performance. One reason can be by utilizing more information, SLT has a better chance to avoid repeated questions. With the help of the user model, the proposed SAMIA framework has the highest reward close to the ground truth.
\begin{table}[ht]
\centering
\caption{The nDCG scores of different models}
\label{tab:nDCG}
\begin{tabular}{c c c c c}
\hline
&{\textbf{SLNT}} & {\textbf{SLT}} & {\textbf{SAMIA}} \\\hline
{nDCG$_1$} & {$0.093\pm0.003$}&{$0.243\pm0.009$}&{$0.383\pm0.017$} \\\hline
{nDCG$_3$} & {$0.158\pm0.009$} &{$0.353\pm0.009$}&{$0.447\pm0.016$}\\\hline
{nDCG$_5$} & {$0.218\pm0.009$} &{$0.374\pm0.011$}&{$0.446\pm0.014$}\\\hline
\end{tabular}
\
\end{table}
\iffalse
\begin{table}[ht]
\centering
\caption{The nDCG scores of different models}
\label{tab:nDCG}
\begin{tabular}{c c c c c}
\hline
&{\textbf{SLNT}} & {\textbf{SLT}} & {\textbf{SAMIA}} \\\hline
{nDCG$_1$} & {0.093}&{0.243}&{0.383} \\\hline
{nDCG$_3$} & {0.158} &{0.353}&{0.447}\\\hline
{nDCG$_5$} & {0.218} &{0.374}&{0.446}\\\hline
\end{tabular}
\
\end{table}
\fi

The nDCG$_p$ scores and their standard errors, where $p$ equals 1, 3 and 5, of all the models on test repeated  five times are shown in Table \ref{tab:nDCG} except the ground truth that has only one response candidate. Obviously, the proposed SAMIA framework significantly outperforms both the SLNT and SLT models in all the settings. This demonstrates that the proposed SAMIA framework not only generates the set of high rewarded candidate responses but also can prioritize the high-reward candidates effectively in the ranking. Due to the extra tag information, the SLT also has some improvement over the vanilla SLNT in all the settings.

\begin{figure}[!htb]
\centering
\includegraphics[scale=0.6]{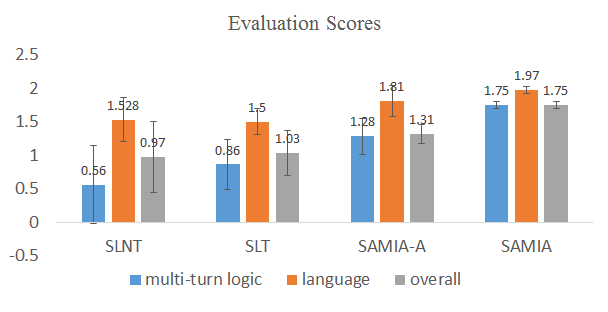}
\caption{Results on human evaluation.}
\label{fig:human}
\end{figure}

As illustrated in Figure~\ref{fig:human}, experimental results on human evaluation show that the SAMIA framework obviously outperforms the others in all 3 aspects, despite that some divergence among human judeges occurs in the SLNT. Both SLNT and SLT mainly suffer from the multi-turn logic performance. Compared with SAMIA-A's score which is around `somewhat suitable' level in the logic and overall quality, the SAMIA framework enhances the robustness and get a score closer to 2. Similar to the previous two criterion-referenced evaluations, the SLT has some improvements in the logic and overall quality compared with SLNT. For the language aspect, all the models have relatively good results with scores larger than 1.5.

\subsubsection{Case Study}

Besides the quantitative evaluation, in Table \ref{tab:example} we conduct a case study to show the effectiveness of the SAMIA framework. In the first two columns, both supervised learning models have the trend to respond the frequent response like the order confirmation but to ignore the missing information when placing orders. The SAMIA-A and SAMIA methods can notice the missing information and ask the user for it. Compared with the SAMIA-A method, the SAMIA framework employing the user model to filter responses can further avoid the universal responses like `OK' and it more actively asks for the missing information to make the dialogue more smooth. Based on this case study, we can see that the SAMIA framework works better than other methods.

\section{Conclusion}

In this paper, we design a novel deep reinforcement learning framework SAMIA to ease the training of a task-oriented dialogue system. There are two steps in the framework. Firstly, we train a user model in a one-turn seq2seq fashion. Secondly, the user model is then adopted to simulate an environment to guide the training of the agent model. In the test phase, the user model also helps filter response candidates to enhance the robustness. The effectiveness of the proposed SAMIA framework has been verified on a real-world coffee ordering dataset. The proposed SAMIA framework has extended deep reinforcement learning to the larger scope and other scenarios, where the modeling of one party in the conversation is easy to be learned by the seq2seq model and the other can be learned by sophisticated models. In the future, we plan to further extend this framework to scenarios with a more complex user modeling and no clear reward.

\newpage

\bibliographystyle{named}
\bibliography{SAMIA}

\begin{thebibliography}{}

\bibitem[\protect\citeauthoryear{Asri \bgroup \em et al.\egroup
  }{2016}]{asri2016sequence}
Layla~El Asri, Jing He, and Kaheer Suleman.
\newblock A sequence-to-sequence model for user simulation in spoken dialogue
  systems.
\newblock {\em arXiv preprint arXiv:1607.00070}, 2016.

\bibitem[\protect\citeauthoryear{Bahdanau \bgroup \em et al.\egroup
  }{2014}]{bahdanau2014neural}
Dzmitry Bahdanau, Kyunghyun Cho, and Yoshua Bengio.
\newblock Neural machine translation by jointly learning to align and
  translate.
\newblock {\em arXiv preprint arXiv:1409.0473}, 2014.

\bibitem[\protect\citeauthoryear{Bordes and Weston}{2016}]{bordes2016learning}
Antoine Bordes and Jason Weston.
\newblock Learning end-to-end goal-oriented dialog.
\newblock {\em arXiv preprint arXiv:1605.07683}, 2016.

\bibitem[\protect\citeauthoryear{Cuay{\'a}huitl}{2016}]{cuayahuitl2016simpleds}
Heriberto Cuay{\'a}huitl.
\newblock Simpleds: A simple deep reinforcement learning dialogue system.
\newblock {\em arXiv preprint arXiv:1601.04574}, 2016.

\bibitem[\protect\citeauthoryear{Dhingra \bgroup \em et al.\egroup
  }{2016}]{dhingra2016end}
Bhuwan Dhingra, Lihong Li, Xiujun Li, Jianfeng Gao, Yun-Nung Chen, Faisal
  Ahmed, and Li~Deng.
\newblock End-to-end reinforcement learning of dialogue agents for information
  access.
\newblock {\em arXiv preprint arXiv:1609.00777}, 2016.

\bibitem[\protect\citeauthoryear{Henderson \bgroup \em et al.\egroup
  }{2014}]{henderson2014second}
Matthew Henderson, Blaise Thomson, and Jason Williams.
\newblock The second dialog state tracking challenge.
\newblock In {\em Proceedings of the 15th Annual Meeting of the Special
  Interest Group on Discourse and Dialogue}, 2014.

\bibitem[\protect\citeauthoryear{Levin \bgroup \em et al.\egroup
  }{1997}]{levin1997learning}
Esther Levin, Roberto Pieraccini, and Wieland Eckert.
\newblock Learning dialogue strategies within the {M}arkov decision process
  framework.
\newblock In {\em Proceedings of 1997 IEEE Workshop on Automatic Speech
  Recognition and Understanding}, pages 72--79, 1997.

\bibitem[\protect\citeauthoryear{Lewis \bgroup \em et al.\egroup
  }{2017}]{DBLP:journals/corr/LewisYDPB17}
Mike Lewis, Denis Yarats, Yann~N. Dauphin, Devi Parikh, and Dhruv Batra.
\newblock Deal or no deal? end-to-end learning for negotiation dialogues.
\newblock {\em CoRR}, abs/1706.05125, 2017.

\bibitem[\protect\citeauthoryear{Li \bgroup \em et al.\egroup
  }{2015}]{li2015diversity}
Jiwei Li, Michel Galley, Chris Brockett, Jianfeng Gao, and Bill Dolan.
\newblock A diversity-promoting objective function for neural conversation
  models.
\newblock {\em arXiv preprint arXiv:1510.03055}, 2015.

\bibitem[\protect\citeauthoryear{Li \bgroup \em et al.\egroup
  }{2016}]{li2016deep}
Jiwei Li, Will Monroe, Alan Ritter, and Dan Jurafsky.
\newblock Deep reinforcement learning for dialogue generation.
\newblock {\em arXiv preprint arXiv:1606.01541}, 2016.

\bibitem[\protect\citeauthoryear{Ranzato \bgroup \em et al.\egroup
  }{2015}]{ranzato2015sequence}
Marc'Aurelio Ranzato, Sumit Chopra, Michael Auli, and Wojciech Zaremba.
\newblock Sequence level training with recurrent neural networks.
\newblock {\em arXiv preprint arXiv:1511.06732}, 2015.

\bibitem[\protect\citeauthoryear{Ritter \bgroup \em et al.\egroup
  }{2011}]{ritter2011data}
Alan Ritter, Colin Cherry, and William~B Dolan.
\newblock Data-driven response generation in social media.
\newblock In {\em Proceedings of the Conference on Empirical Methods in Natural
  Language Processing}, pages 583--593, 2011.

\bibitem[\protect\citeauthoryear{Rush \bgroup \em et al.\egroup
  }{2015}]{rush2015neural}
Alexander~M Rush, Sumit Chopra, and Jason Weston.
\newblock A neural attention model for abstractive sentence summarization.
\newblock {\em arXiv preprint arXiv:1509.00685}, 2015.

\bibitem[\protect\citeauthoryear{Schatzmann \bgroup \em et al.\egroup
  }{2007}]{schatzmann2007statistical}
Jost Schatzmann, Blaise Thomson, and Steve Young.
\newblock Statistical user simulation with a hidden agenda.
\newblock In {\em Proceedings of the 8th SIGDial Workshop on Discourse and
  Dialogue}, 2007.

\bibitem[\protect\citeauthoryear{Serban \bgroup \em et al.\egroup
  }{2015}]{serban2015hierarchical}
Iulian~V Serban, Alessandro Sordoni, Yoshua Bengio, Aaron Courville, and Joelle
  Pineau.
\newblock Hierarchical neural network generative models for movie dialogues.
\newblock {\em arXiv preprint arXiv:1507.04808}, 2015.

\bibitem[\protect\citeauthoryear{Shang \bgroup \em et al.\egroup
  }{2015}]{shang2015neural}
Lifeng Shang, Zhengdong Lu, and Hang Li.
\newblock Neural responding machine for short-text conversation.
\newblock {\em arXiv preprint arXiv:1503.02364}, 2015.

\bibitem[\protect\citeauthoryear{Sordoni \bgroup \em et al.\egroup
  }{2015}]{sordoni2015hierarchical}
Alessandro Sordoni, Yoshua Bengio, Hossein Vahabi, Christina Lioma, Jakob
  Grue~Simonsen, and Jian-Yun Nie.
\newblock A hierarchical recurrent encoder-decoder for generative context-aware
  query suggestion.
\newblock In {\em Proceedings of the 24th ACM International on Conference on
  Information and Knowledge Management}, pages 553--562, 2015.

\bibitem[\protect\citeauthoryear{Su \bgroup \em et al.\egroup
  }{2016}]{su2016line}
Pei-Hao Su, Milica Gasic, Nikola Mrksic, Lina Rojas-Barahona, Stefan Ultes,
  David Vandyke, Tsung-Hsien Wen, and Steve Young.
\newblock On-line active reward learning for policy optimisation in spoken
  dialogue systems.
\newblock {\em arXiv preprint arXiv:1605.07669}, 2016.

\bibitem[\protect\citeauthoryear{Sutskever \bgroup \em et al.\egroup
  }{2014}]{sutskever2014sequence}
Ilya Sutskever, Oriol Vinyals, and Quoc~V Le.
\newblock Sequence to sequence learning with neural networks.
\newblock In {\em Advances in Neural Information Processing Systems 27}, pages
  3104--3112, 2014.

\bibitem[\protect\citeauthoryear{Walker \bgroup \em et al.\egroup
  }{2003}]{walker2003trainable}
Marilyn~A Walker, Rashmi Prasad, and Amanda Stent.
\newblock A trainable generator for recommendations in multimodal dialog.
\newblock In {\em INTERSPEECH}, 2003.

\bibitem[\protect\citeauthoryear{Wen \bgroup \em et al.\egroup
  }{2015}]{DBLP:conf/emnlp/WenGMSVY15}
Tsung{-}Hsien Wen, Milica Gasic, Nikola Mrksic, Pei{-}hao Su, David Vandyke,
  and Steve~J. Young.
\newblock Semantically conditioned lstm-based natural language generation for
  spoken dialogue systems.
\newblock In {\em Proceedings of the 2015 Conference on Empirical Methods in
  Natural Language Processing, {EMNLP} 2015, Lisbon, Portugal, September 17-21,
  2015}, pages 1711--1721, 2015.

\bibitem[\protect\citeauthoryear{Wen \bgroup \em et al.\egroup
  }{2016}]{wen2016network}
Tsung-Hsien Wen, Milica Gasic, Nikola Mrksic, Lina~M Rojas-Barahona, Pei-Hao
  Su, Stefan Ultes, David Vandyke, and Steve Young.
\newblock A network-based end-to-end trainable task-oriented dialogue system.
\newblock {\em arXiv preprint arXiv:1604.04562}, 2016.

\bibitem[\protect\citeauthoryear{Williams \bgroup \em et al.\egroup
  }{2017}]{DBLP:conf/acl/WilliamsAZ17}
Jason~D. Williams, Kavosh Asadi, and Geoffrey Zweig.
\newblock Hybrid code networks: practical and efficient end-to-end dialog
  control with supervised and reinforcement learning.
\newblock In {\em Proceedings of the 55th Annual Meeting of the Association for
  Computational Linguistics, {ACL} 2017, Vancouver, Canada, July 30 - August 4,
  Volume 1: Long Papers}, pages 665--677, 2017.

\bibitem[\protect\citeauthoryear{Williams}{1992}]{williams1992simple}
Ronald~J Williams.
\newblock Simple statistical gradient-following algorithms for connectionist
  reinforcement learning.
\newblock {\em Machine learning}, 8(3-4):229--256, 1992.

\bibitem[\protect\citeauthoryear{Young \bgroup \em et al.\egroup
  }{2010}]{young2010hidden}
Steve Young, Milica Ga{\v{s}}i{\'c}, Simon Keizer, Fran{\c{c}}ois Mairesse,
  Jost Schatzmann, Blaise Thomson, and Kai Yu.
\newblock The hidden information state model: A practical framework for
  pomdp-based spoken dialogue management.
\newblock {\em Computer Speech \& Language}, 24(2):150--174, 2010.

\bibitem[\protect\citeauthoryear{Young \bgroup \em et al.\egroup
  }{2013}]{young2013pomdp}
Steve Young, Milica Ga{\v{s}}i{\'c}, Blaise Thomson, and Jason~D Williams.
\newblock Pomdp-based statistical spoken dialog systems: A review.
\newblock {\em Proceedings of the IEEE}, 101(5):1160--1179, 2013.

\end{thebibliography}

\end{document}